\documentclass[runningheads]{llncs}
\usepackage{graphicx}

\usepackage{tikz}
\usepackage{comment}
\usepackage{amsmath,amssymb} 
\usepackage{color}
\usepackage{subcaption}
\usepackage{booktabs}
\usepackage{xurl}

\usepackage[accsupp]{axessibility}  

\begin{document}
\pagestyle{headings}
\mainmatter
\def\ECCVSubNumber{8}  

\title{Abstract Images Have Different Levels of Retrievability Per Reverse Image Search Engine} 


\titlerunning{Abstract Images Have Different Levels of Retrievability}
%
\author{Shawn M. Jones\inst{1}\orcidID{0000-0002-4372-870X}\index{Jones, Shawn M.} \and
Diane Oyen\inst{1}\orcidID{0000-0002-1353-3688}\index{Oyen, Diane}}
\authorrunning{S. Jones \& D. Oyen}
%
\institute{Los Alamos National Laboratory, Los Alamos NM 87545, USA
\email{\{smjones,doyen\}@lanl.gov}}
\maketitle

\begin{abstract}
Much computer vision research has focused on natural images, but technical documents typically consist of abstract images, such as charts, drawings, diagrams, and schematics. How well do general web search engines discover abstract images? Recent advancements in computer vision and machine learning have led to the rise of reverse image search engines. Where conventional search engines accept a text query and return a set of document results, including images, a reverse image search accepts an image as a query and returns a set of images as results. This paper evaluates how well common reverse image search engines discover abstract images. We conducted an experiment leveraging images from Wikimedia Commons, a website known to be well indexed by Baidu, Bing, Google, and Yandex. We measure how difficult an image is to find again (retrievability), what percentage of images returned are relevant (precision), and the average number of results a visitor must review before finding the submitted image (mean reciprocal rank). When trying to discover the same image again among similar images, Yandex performs best. When searching for pages containing a specific image, Google and Yandex outperform the others when discovering photographs with precision scores ranging from 0.8191 to 0.8297, respectively. In both of these cases, Google and Yandex perform better with natural images than with abstract ones achieving a difference in retrievability as high as 54\% between images in these categories. These results affect anyone applying common web search engines to search for technical documents that use abstract images.
\end{abstract}

\begin{figure}

    \begin{subfigure}{\textwidth}
        \centering
        \includegraphics[width=\textwidth]{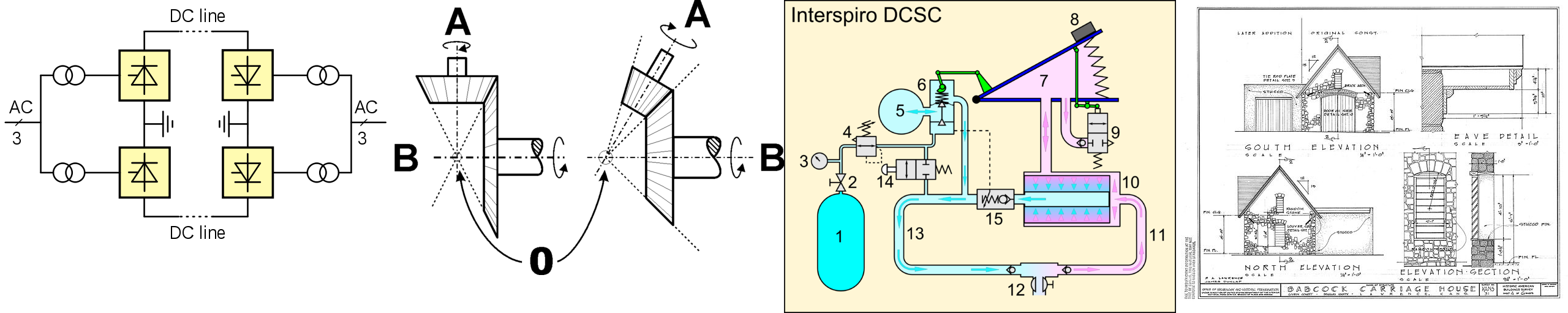}
        \caption*{\centering from left: Bipolar Hi Voltage DC, Bevel Gear, diving rebreather, carriage house}
        \caption{schematic}
        \label{fig:schematic-example}
    \end{subfigure}
    \par\bigskip
    \begin{subfigure}{\textwidth}
        \centering
        \includegraphics[width=\textwidth]{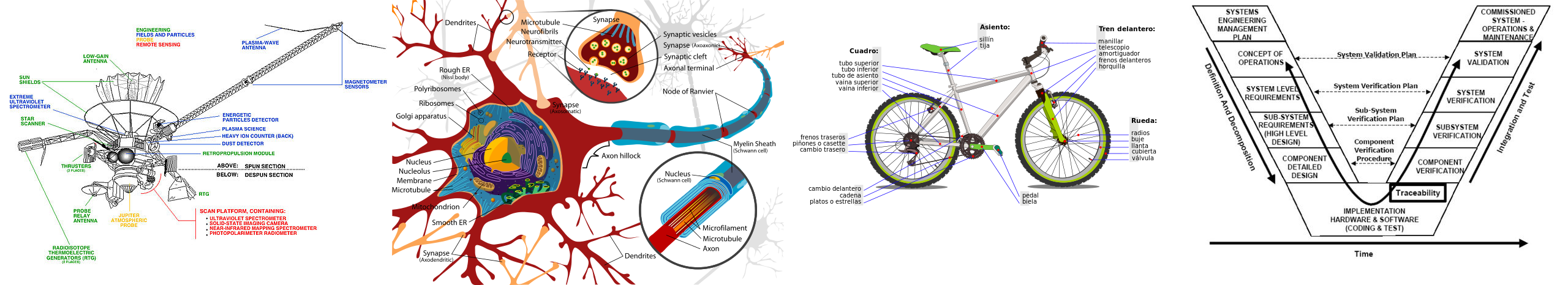}
        \caption*{\centering from left: Galileo spacecraft, neuron, bicycle, systems engineering V diagram}
        \caption{\centering diagram}
        \label{fig:diagram-example}
    \end{subfigure}
    \par\bigskip
    \begin{subfigure}{\textwidth}
    \centering
        \includegraphics[height=0.2\textheight]{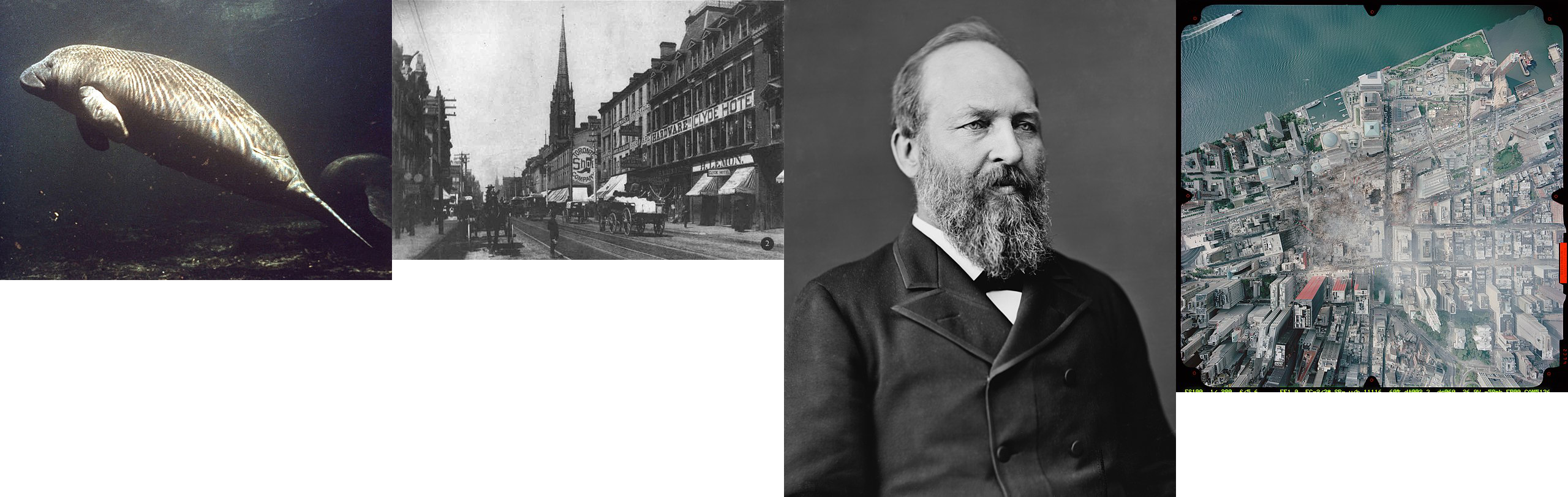}
        \caption*{\centering from left: manatee, downtown Toronto 1890, James Garfield, \newline World Trade Center ground zero}
        \caption{\centering photo}
        \label{fig:photo-example}
    \end{subfigure}
    \par\bigskip
    \begin{subfigure}{\textwidth}
    \centering
        \includegraphics[width=\textwidth]{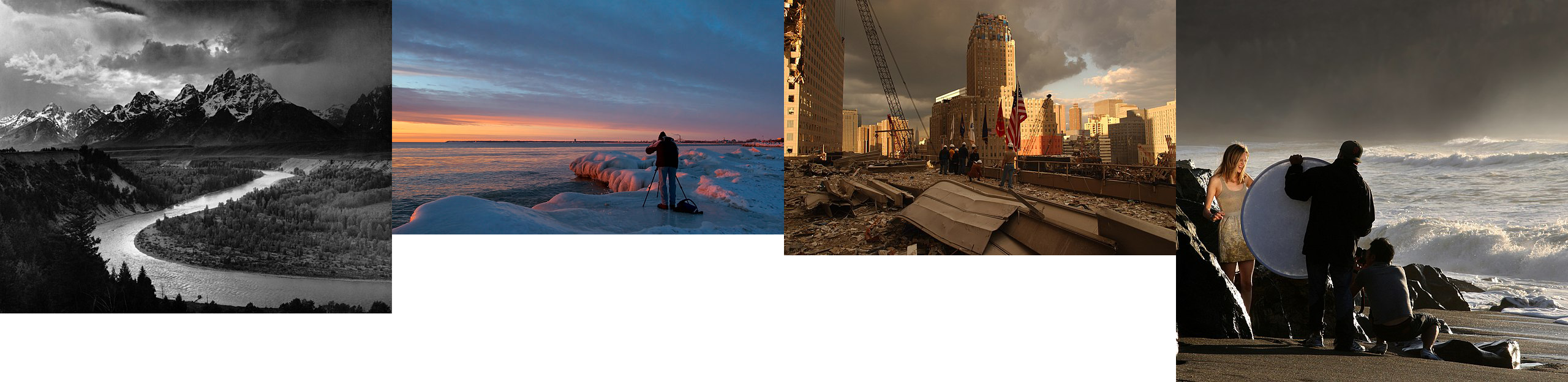}
        \caption*{\centering from left: the Snake River, sunrise over Lake Michigan, New York City, photographing a model}
        \caption{photograph}
        \label{fig:photograph-example}
    \end{subfigure}

   \caption{Example images from Wikipedia Commons of our image categories. The categories of \emph{schematic} and \emph{diagram} represent abstract images while \emph{photo} and \emph{photograph} represent natural ones.}
   \label{fig:example-query-images}
\end{figure}

\begin{figure}

    \begin{subfigure}{0.5\textwidth}
        \includegraphics[width=\textwidth]{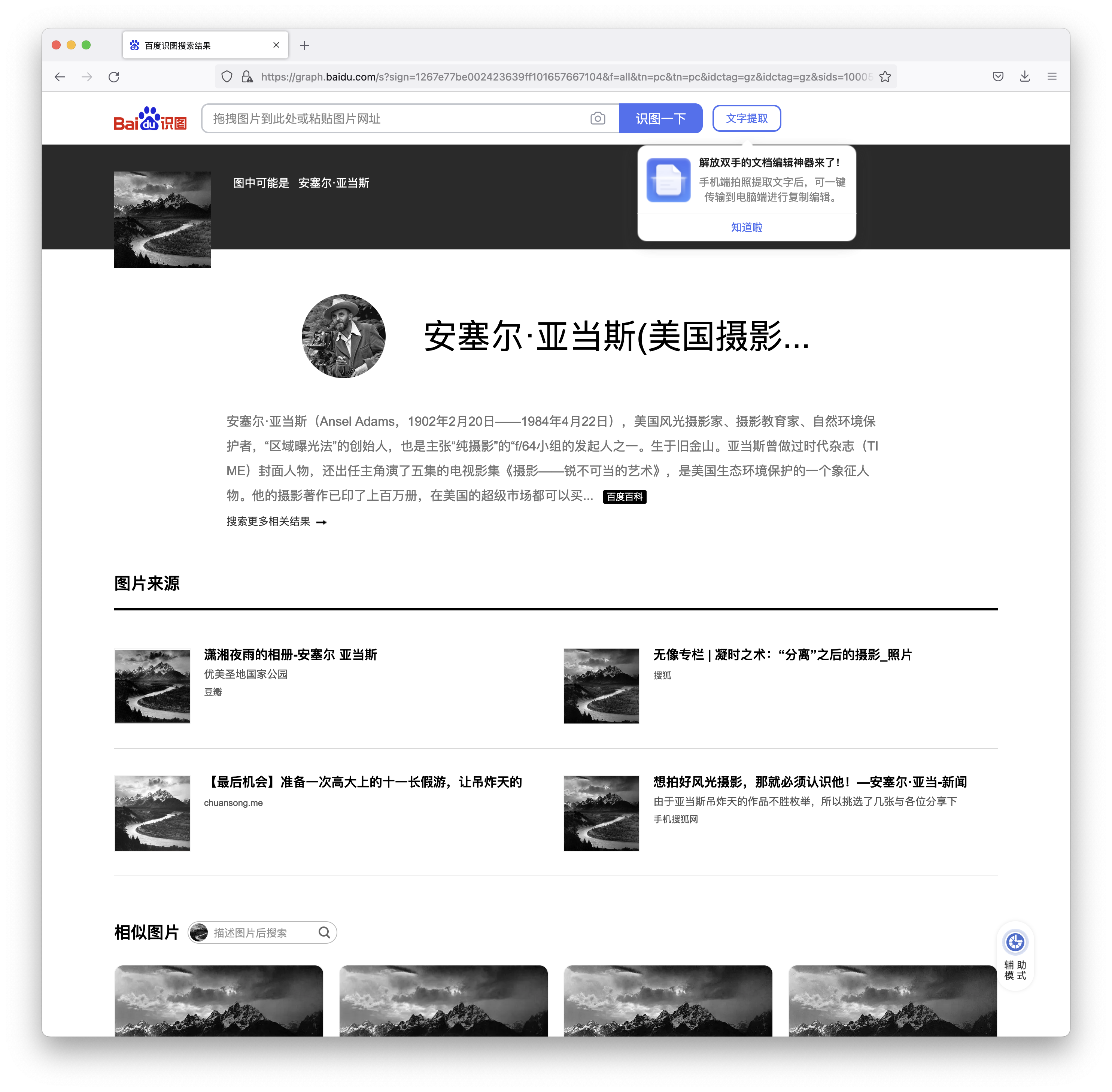}
        \caption{Baidu}
        \label{fig:baidu}
    \end{subfigure}
    \hfill
    \begin{subfigure}{0.5\textwidth}
        \includegraphics[width=\textwidth]{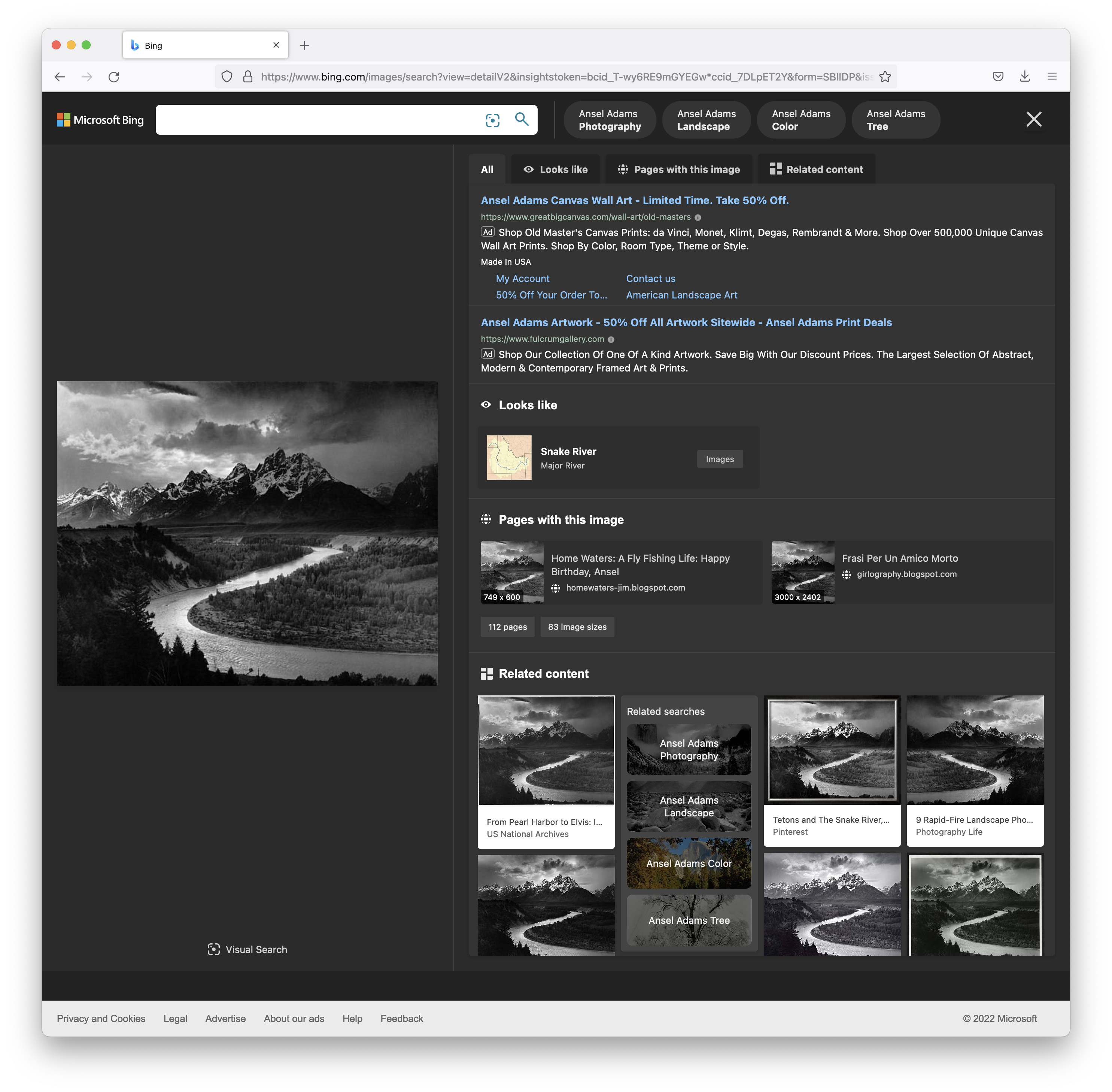}
        \caption{Bing}
        \label{fig:bing}
    \end{subfigure}
    \hfill
    \begin{subfigure}{0.5\textwidth}
        \includegraphics[width=\textwidth]{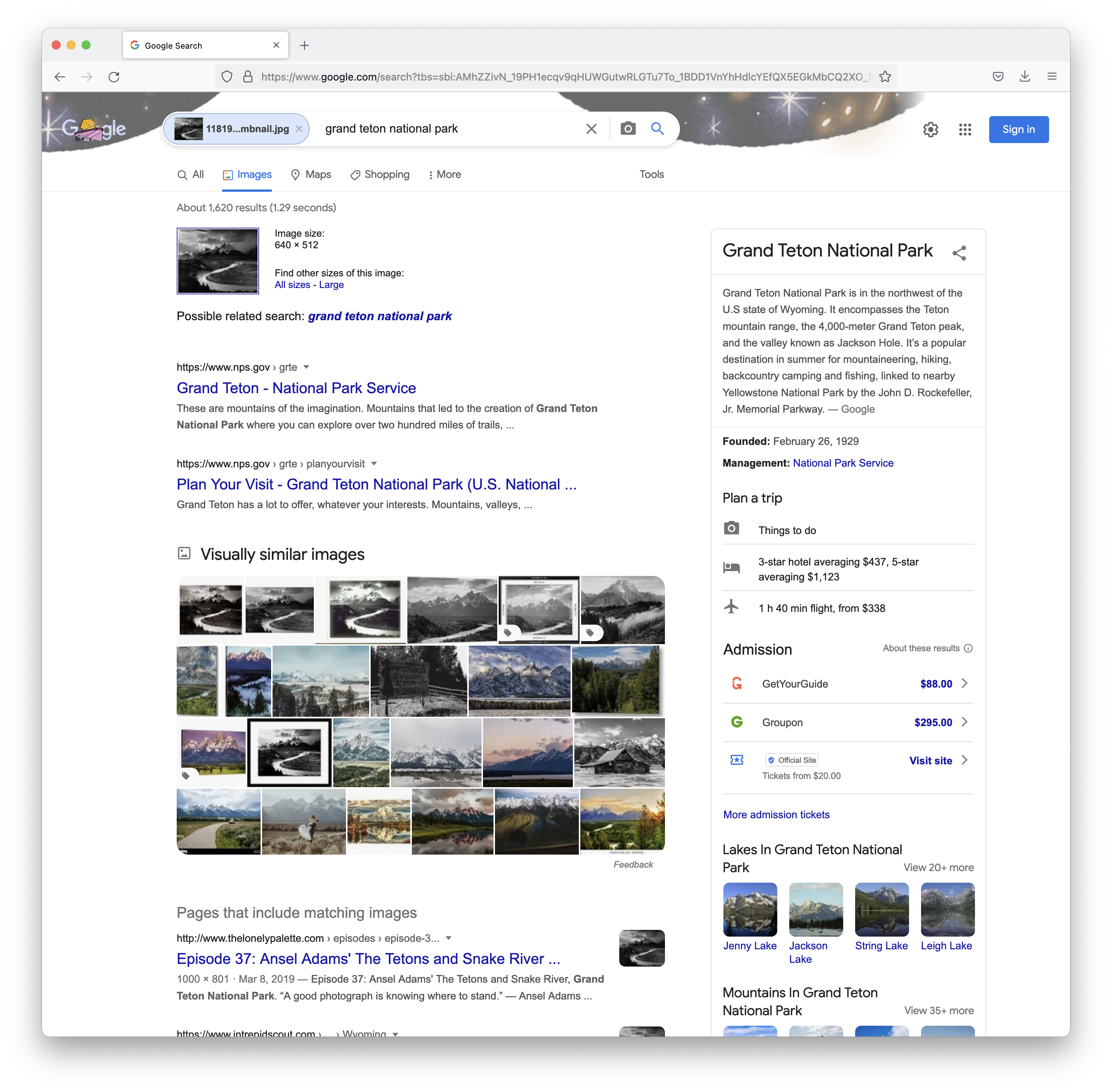}
        \caption{Google}
        \label{fig:google}
    \end{subfigure}
    \hfill
    \begin{subfigure}{0.5\textwidth}
        \includegraphics[width=\textwidth]{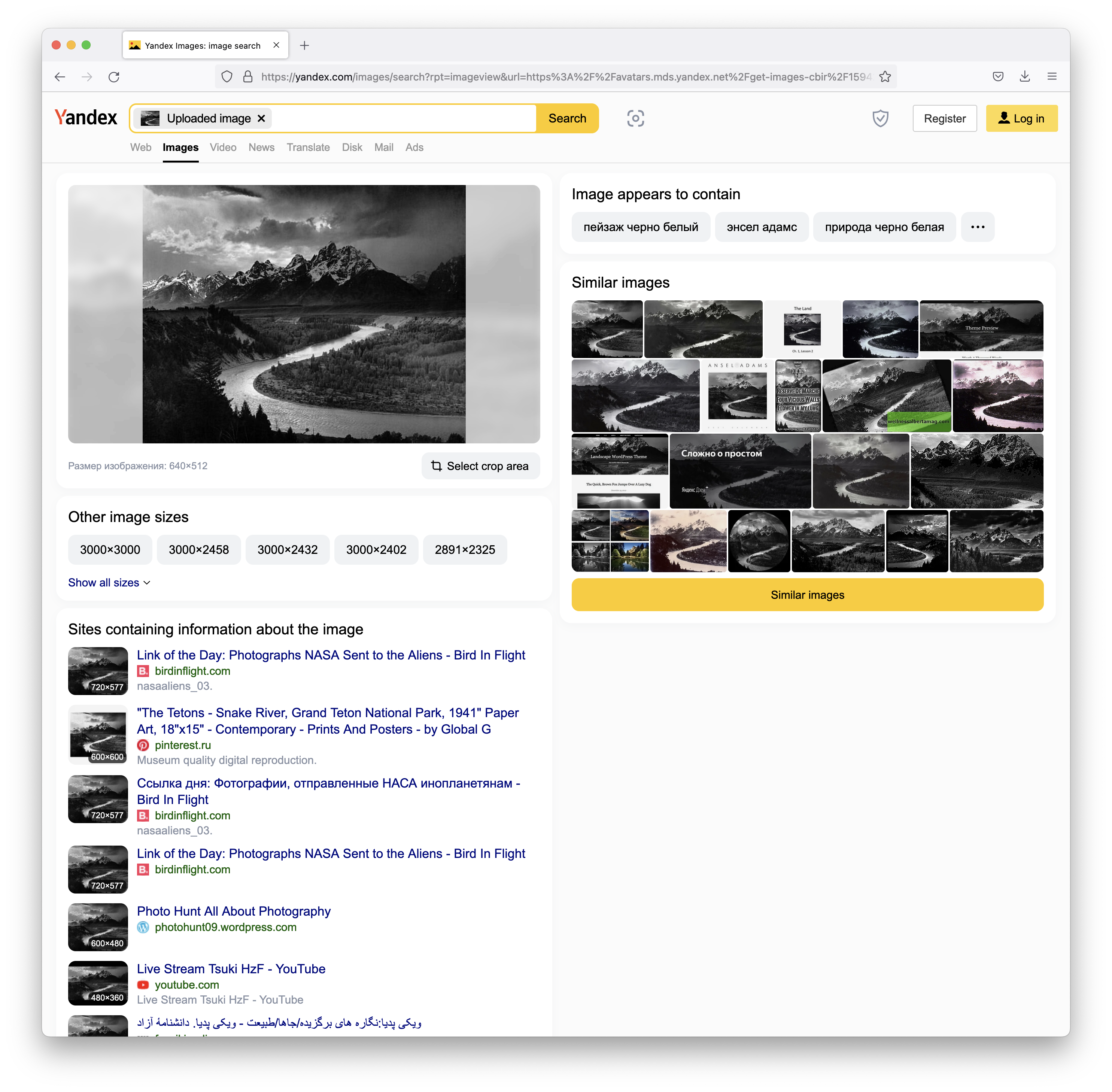}
        \caption{Yandex}
        \label{fig:yandex}
    \end{subfigure}

   \caption{Baidu, Bing, Google, and Yandex support similar outputs if a visitor uploads an image file. A visitor can choose to view images similar to the input, or view the pages with the image that they uploaded.}
   \label{fig:both-search-engines}
\end{figure}

\section{Introduction}

Technical documents, such as manuals and research papers, contain visualizations in the form of abstract images such as charts, diagrams, schematics, and illustrations. Yet, technical information is often searched and retrieved by text alone \cite{caragea2014citeseer,clark2016pdffigures,li2018deeppatent}. Meanwhile, computer vision has significantly advanced the ability to search and retrieve images \cite{gordo2017end,radenovic2018fine}, including sketch-based retrieval \cite{Ribeiro_2020_CVPR,sketchy2016,song2017fine,xu2022deep}. Yet there are few computer vision research papers focused on querying and retrieving abstract, technical drawings \cite{Kucer_2022_WACV,piroi2011clef,vrochidis2012concept}, and so it is unclear whether the computer vision approaches that are successful with general images are similarly successful with technical drawings. Rather than evaluating computer vision algorithms directly, we used common search engines to evaluate whether these black-box systems (presumably incorporating state-of-the-art image retrieval algorithms) work as well in retrieving diagrams as they do with retrieving photographs. Our goal is to highlight the gap between the success of searching general images on the web and the capability to search images containing technical information on the web. We achieve this goal through the development of a method to quantify the retrievability of schematics (Figure \ref{fig:schematic-example}) and diagrams (Figure \ref{fig:diagram-example})  -- abstract images -- versus photos (Figure \ref{fig:photo-example}) and photographs (Figure \ref{fig:photograph-example}) -- natural images -- using common search engines. In this work, we provide an answer to the following research question: 

\emph{When using the reverse image search capability of general web search engines, are natural images more easily discovered than abstract ones?}

Our experiment uses the \textbf{reverse image search} capability of Baidu, Bing, Google, and Yandex. Rather than submitting textual queries and receiving images as \textbf{search engine results (SERs)}, with reverse image search, we upload image files. As seen in Figure \ref{fig:both-search-engines}, we uploaded an image of the Snake River to all four search engines. From here, a visitor can choose to explore images that are \textbf{similar-to} their uploaded image, or they can also view the list of \textbf{pages-with} the uploaded image, including images that are highly visually similar.

We used Wikimedia Commons as a source of images to upload. Prior work has focused on the presence of \emph{Wikipedia} content in Google SERs \cite{mcmahon_substantial_2017} as well as those from DuckDuckGo and Bing \cite{vincent_deeper_2021}. These efforts established that Wikipedia articles have high retrievability for certain types of queries. Search engines discover images through their source documents. Thus we assume that the images in Wikipedia documents are equally retrievable.

We experimented with these four search engines. We acquired 200 abstract images from Wikimedia Commons with the search terms \emph{diagram} and \emph{schematic} and 199 natural images with the search terms \emph{photo} and \emph{photograph}. We submitted these images to the reverse image search engines at Baidu, Bing, Google, and Yandex. From there, we evaluated how often the search engine returned the same image in the results, as established by perceptual hash \cite{krawetz_2011,10.1117/12.2594720,zauner_implementation_2010}. We evaluated the SERs in terms of precision, mean reciprocal rank \cite{voorhees_trec_1999}, and retrievability \cite{azzopardi_retrievability_2008}. Yandex performed best in all cases. When searching for pages containing a specific image, Google and Yandex outperform the others when discovering photographs with precision scores ranging from 0.8191 to 0.8297, respectively. In both of these cases, Google and Yandex perform better with natural images than with abstract ones achieving a difference in retrievability as high as 54\% between images in these categories. These results indicate a clear difference in capability between natural images and abstract images, which will affect a technical document reader's ability to complete the task of discovering similar images or other documents containing an image of interest.

\section{Background}

Our study primarily focuses on using image files as queries, which we refer to as \textbf{query images}. Using image files as queries makes our work different from many other studies evaluating public search engines. Similar constructs exist within the SERs at the search engines in this study. As noted above, these four search engines support \textbf{similar-to} and \textbf{pages-with} SERs. The screenshots in Figure \ref{fig:both-search-engines} demonstrate how a visitor is presented with summaries of these options before they choose the type of SER they would like to explore further. Each SER contains an \textbf{image URL} that identifies the image indexed by the search engine. Each SER also provides a \textbf{page URL} identifying the page where a visitor can find that image in use. If the image from the image URL is no longer available, each SER also contains a \textbf{thumbnail URL} that represents a smaller version of the image. The search engine hosts these thumbnails to speed up the presentation of results.

\subsection{Metrics to Evaluate Search Engines}

\begin{equation}
   Precision@k = \begin{cases}
        \frac{| \textrm{relevant SERs} \, \cap \, \textrm{retrieved SERs} |}{k} \quad & \textrm{if} \, k < |\textrm{retrieved SERs}| \\
        \frac{| \textrm{relevant SERs} |}{| \textrm{retrieved SERs} |} \quad & \textrm{otherwise} \\
   \end{cases} 
   \label{eq:precision}
\end{equation}

We apply different metrics to measure the performance of the different Reverse Image Search engines. Equation \ref{eq:precision} demonstrates how to calculate \textbf{precision} at a cutoff $k$, also written \textbf{$p@k$}. This measure helps us understand the performance of the search engine by answering the question ``What is the percentage of images in the SERs that are relevant if we stop at $k$ results?'' If the most relevant results exist at the beginning of a set of SERs, then precision decreases as the value of $k$ increases.

\begin{equation}
   r(d) = \sum_{q \in Q} o_q \cdot f(k_{dq}, c)
   \label{eq:retrievability}
\end{equation}

Second, we want to understand if the search engine returned the query image within a given set of results. Equation \ref{eq:retrievability} shows how to calculate Azzopardi's retrievability \cite{azzopardi_retrievability_2008}. Retrievability helps us answer the question, ``Given a document, was it retrieved within the cutoff $c$?'' Retrievability leverages a cutoff $c$, similar to $p@k$'s $k$, to indicate how many results the user reviews before stopping. Its cost function $f(k_{dq}, c)$ returns 1 if document $d$ is found within cutoff $c$ for query $q$, and 0 otherwise. As we increase $c$, we increase the chance of the search engine returning a relevant document; hence retrievability increases.

\begin{equation}
   MRR = \frac{1}{Q} \sum^{|Q|}_{i=1} \frac{1}{rank_i}
   \label{eq:mrr}
\end{equation}

Finally, we measure a visitor's effort before finding their query image in the results. Equation \ref{eq:mrr} demonstrates how to calculate \textbf{mean reciprocal rank} (\textbf{MRR}). MRR helps us answer the question ``How many results, on average across all queries, must a visitor review before finding a relevant one?'' For example, if the first result is relevant for an individual query, its reciprocal rank is $\frac{1}{1}$. Its reciprocal rank is $\frac{1}{2}$ if the second is relevant. If the third is relevant, it is $\frac{1}{3}$. MRR is the mean of these reciprocal ranks across all queries. An MRR of 1.0 indicates that the first result was relevant for all queries. An MRR of 0 indicates that no queries returned a relevant result.

\subsection{Similarity Approaches To Establish Relevance}

Each of these metrics requires that we identify which SERs are relevant. In our case, ``relevant'' means that it is the same image we uploaded. To determine if an image is relevant, we apply a perceptual hash \cite{zauner_implementation_2010} to each image. This perceptual hash provides a non-unique hash describing the content of the image. We compute the distances between two hashes to determine if two images are the same. If they score below some threshold, then we consider them the same. Perceptual hashes are intended to be a stand-in for human evaluation of the similarity of two images. Many perceptual hash algorithms exist. For this work, we apply ImageHash's \cite{imagehash} implementation of pHash \cite{krawetz_2011} and GoFigure's \cite{gofigure} implementation of VisHash \cite{10.1117/12.2594720}.

ImageHash's pHash, hereafter referred to as \textbf{pHash}, was designed to compare photographs. It applies Discrete Cosine Transforms (DCT). A user compares pHashes by computing the Hamming distance between them. As established by Krawetz \cite{krawetz_2011} we use a distance of 5-bits to indicate that two images are likely the same. pHash \cite{krawetz_2011,zauner_implementation_2010} demonstrates improvements over simpler algorithms such as the average hash algorithm (\textbf{aHash}), which computes a hash based on the mean of color values in each pixel. Fei et al. \cite{FEI2015413}, Chamoso \cite{10.1007/978-3-319-61578-3_18}, Arefkhani \cite{7397539} and Vega \cite{8226396} have demonstrated that pHash is more accurate than aHash in a variety of applications. These studies also evaluated the Difference hashing \cite{krawetz_2013} (\textbf{dHash}) algorithm and found that it performs almost as well as pHash.

\textbf{VisHash} \cite{10.1117/12.2594720} was designed as an extension to dHash with a focus on providing a similarity metric for diagrams and technical drawings. VisHash relies more heavily on the shapes in the image, whereas pHash relies on frequencies. A user compares VisHashes by computing a normalized L2 distance between them. As established by Oyen et al. \cite{10.1117/12.2594720}, we use a distance of $\le 0.3$ to determine if a query image is the same as the one in its SER. VisHash has fewer collisions than pHash and fewer false positive matches, especially for diagrams and technical drawings.

\section{Related Work}

Precision and MRR \cite{croft_information_2015,manning_2008} are well-known metrics applied in evaluating search engines, especially with the established TREC dataset \cite{voorhees2005trec}. Retrievability is the newer metric. Azzopardi conducted several studies \cite{10.1145/2808194.2809444,10.1007/978-3-319-06028-6_85,azzopardi_retrievability_2008,10.1145/3132847.3133151,10.1145/2484028.2484145,10.1007/978-3-319-06028-6_82,10.1145/2661829.2661948,10.1007/978-3-319-16354-3_22,10.1145/3132847.3133135} to support the effectiveness of the retrievability metric with different search engines and corpora. Retrievability has been used to evaluate simulated user queries \cite{traub_bias_2016}, assess the impact of OCR errors on information retrieval \cite{vanstrien_2020}, examine retrievability in web archives \cite{jones_dissertation,samar_web_archive_2018}, and expose issues with retrievability of patents and legal documents \cite{10.1145/1645953.1646250}. These studies give us confidence in this metric's capability. As far as we know, we are the first to apply it to image SERs.

In a 2013 manual evaluation, Nieuwenhuysen \cite{nieuwenhuysen2013search} found Google's image search to have higher precision than competitor Tineye\footnote{\url{https://tineye.com/}}. They repeated their manual study in 2018 \cite{nieuwenhuysen2019finding}, and included Yandex but still discovered that Google performed best. In 2015, Kelly \cite{kelly_2015} conducted a similar study and found both Google and Tineye performed poorly. In 2020, Bitirim et al. \cite{9202716} manually evaluated Google's reverse image search capability using 25 query images and declared its precision poor (52\%) across multiple categories of items. Our study differs because we use 380 query images and use perceptive hashes to evaluate similarity, thus avoiding human disagreements about relevance. We also include both types of SER results.

d'Andr{\'e}a and Mintz \cite{dandrea_2018,IJoC10423} noted that reverse image searches could help researchers study events for which they do not speak the language but wish to find related sources. Similar studies were conducted on images about repealing Ireland's 8th Amendment \cite{curran_2022}, Dutch politics \cite{zahorova_2018}, biological research \cite{https://doi.org/10.1002/0471142727.mb1913s111}, disinformation/misinformation \cite{10.1007/978-3-031-09680-8_25,mci/Askinadze2017,dhanvi_2022}, dermatologic diagnoses \cite{SHARIFZADEH2021202,JIA20211415,10.1001/jamadermatol.2016.2096}, and the spread of publicly available images \cite{https://doi.org/10.1002/asi.23847}. Our work differs because it evaluates the search engines themselves for retrievability rather than as a tool to address another research question.

Horv{\'a}th \cite{10.1117/12.2228505} applied reverse image search to construct a dataset useful for training classifiers. Similarly, Guinness et al. \cite{10.1145/3173574.3174092} performed reverse image searches with Google and scraped captions for the visually impaired. Chutel and Sakare \cite{6950085,chutel2014reverse} summarized existing reverse image search techniques. Gaillard and Egyed-Zsigmond \cite{gaillard2017large,gaillard2018}, Araujo et al. \cite{ARAUJO201835}, Mawoneke et al. \cite{9339350}, Diyasa et al. \cite{9321037}, Veres et al. \cite{veres2018choosing}, Gandhi et al. \cite{gandhireverse} evaluated their own reverse image search algorithms. Rather than evaluating our own system,  we evaluate how well the state of the art publicly-available reverse image search engines function for discovering the same image again. We are more interested in whether the type of image matters to the results.

Perceptual hashes have been applied to verify images converted for digital preservation \cite{heritage2020075}, to detect plagiarism \cite{10.1145/3197026.3197042}, to evaluate machine learning results \cite{jones_automatically_selecting_striking_2021}, and to uncover disinformation \cite{Zannettou_2020}. Alternatives to our chosen pHash and VisHash methods include work by Ruchay et al. \cite{10.1117/12.2272716}, Monga and Evans \cite{1709989}, Cao et al. \cite{8003435}, and Lei et al. \cite{LEI2011280}. We chose ImageHash's pHash for this preliminary study due to its prevalence in other literature and its robust implementation as part of ImageHash. We selected GoFigure's VisHash as an alternative for comparison because it was specifically designed for drawings and diagrams.

\section{Methods}

To conduct this experiment, we needed a set of query images that we could reasonably be assured were indexed by search engines, so we turned to Wikimedia Commons. We wanted images from the categories of \emph{diagram} and \emph{schematic} (abstract imagery) as well as \emph{photo} and \emph{photograph} (natural images), so we submitted these search terms to the Wikimedia Commons API in January 2022 and recorded the resultant images. Figure \ref{fig:example-query-images} shows example Wikimedia Commons images from each of these queries. Those returned from the search terms \emph{diagram} or \emph{schematic} reflect abstract imagery. Those returned from the search terms \emph{photo} and \emph{photograph} represent natural images. We took the first 100 images returned from each category. One image failed to download. Two images overlap between the \emph{schematic} and \emph{diagram} categories. Seventeen images overlap between the \emph{photo} and \emph{photograph} categories. Of the 400 Wikimedia Commons images returned by the API, we were left with 380 unique images to use as queries for the reverse image search engines.

Because the images have many different sizes and search engines have file size limitations for uploaded images, we asked the Wikimedia API to generate images that were 640px wide, scaling their height to preserve their aspect ratio.

\begin{table}
\caption{The number of queries for which we acquired \emph{similar-to} results}
\label{tab:query-counts-similar-to}
\centering
\begin{tabular}{l r @{\hskip .5cm} r @{\hskip .5cm} r @{\hskip .5cm} r @{\hskip .5cm} || r}
\toprule
            &   Baidu &   Bing &   Google &   Yandex & Category \\
            &         &        &          &          & Total \\
\midrule
 diagram    &      82 &     83 &       79 &       97 & 100 \\
 schematic  &      74 &     87 &       82 &      100 & 100 \\
 photo      &      80 &     78 &       72 &       93 & 100 \\
 photograph &      77 &     85 &       79 &       94 & 99 \\
\midrule
 Total     & 313 & 333 & 310 & 384 & 399 \\
 & & & & & (380 unique) \\
\bottomrule
\end{tabular}

\end{table}

\begin{table}
\caption{The number of queries for which we acquired \emph{pages-with} results}
\label{tab:query-counts-pages-with}
\centering
\begin{tabular}{l r @{\hskip .5cm} r @{\hskip .5cm} r @{\hskip .5cm} r @{\hskip .5cm} || r}
\toprule
            &   Baidu &   Bing &   Google &   Yandex & Category \\
            &         &        &          &          & Total \\
\midrule
 diagram    &      64 &     79 &       97 &       94 & 100 \\
 schematic  &      54 &     57 &       87 &       95 & 100 \\
 photo      &      38 &     44 &       71 &       94 & 100 \\
 photograph &      38 &     50 &       76 &       94 & 99 \\
\midrule
 Total      &    194  & 230  & 331 & 377 & 399 \\
 & & & & & (380 unique) \\
\bottomrule
\end{tabular}

\end{table}

We developed scraping programs to submit images and extract both pages-with and similar-to SERs from Baidu, Bing, Google, and Yandex. To provide the best evaluation, we removed any context from the image files that might impart additional metadata to the search engine. To avoid the possibility that the search engines might apply the filename to its analysis, we changed the file name of each image to \texttt{upload\_file.ext} where \texttt{ext} reflected the appropriate extension for the content type. We opted to use each search engine's file upload feature rather than providing the URLs directly from Wikimedia to ensure that the URL did not influence the results. We ran the scraping programs in June 2022 to successfully acquire the number of results shown in Tables \ref{tab:query-counts-similar-to} and \ref{tab:query-counts-pages-with}.

We submitted the same 380 query images to each search engine's scraping program and captured the pages-with and similar-to results. We recorded the SER URL, page URL, image URL, thumbnail URL, and position of a maximum of 100 SERs from each search engine. In some cases, we did not have 100 SERs. Because downloading image URLs resulted in some images that were no longer available, we downloaded all 154,191 thumbnail images because they were more reliable. Thumbnail sizes differ by search engine, ranging in size from 61px wide with Google's page-with results to 480px tall with Yandex's similar-to results. All of our metrics require a relevance determination for each result. Due to the scale, rather than applying human judgment, we considered an SER relevant if the distance between its thumbnail image and the query image fell under the VisHash or pHash thresholds mentioned earlier.

We encountered issues downloading 8,268/154,191 (5.362\%) thumbnail images. VisHash encountered issues processing 170/154,191 (0.110\%) thumbnail images. Because these results could not be analyzed, we treated them as if they were not relevant for metrics purposes.

\section{Results and Discussion}

\begin{figure}
    \centering
    \includegraphics[width=\textwidth]{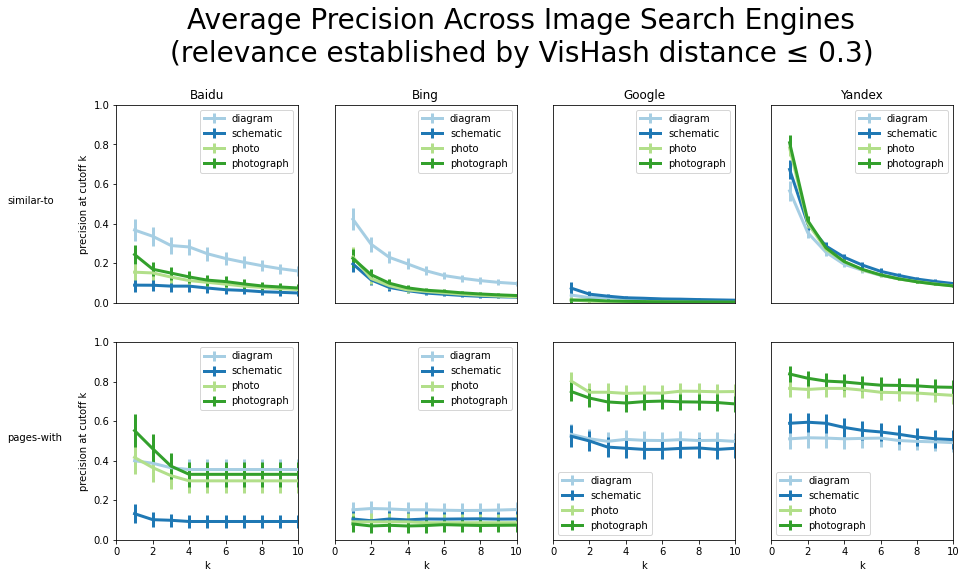}
    \caption{Average precision at cutoff $k$ between the values of 1 and 10 for Wikimedia Commons query images, using a VisHash threshold of 0.3. Error bars represent standard error at that value of $k$.}
    \label{fig:precision-vishash}
\end{figure}

\begin{figure}
    \centering
    \includegraphics[width=\textwidth]{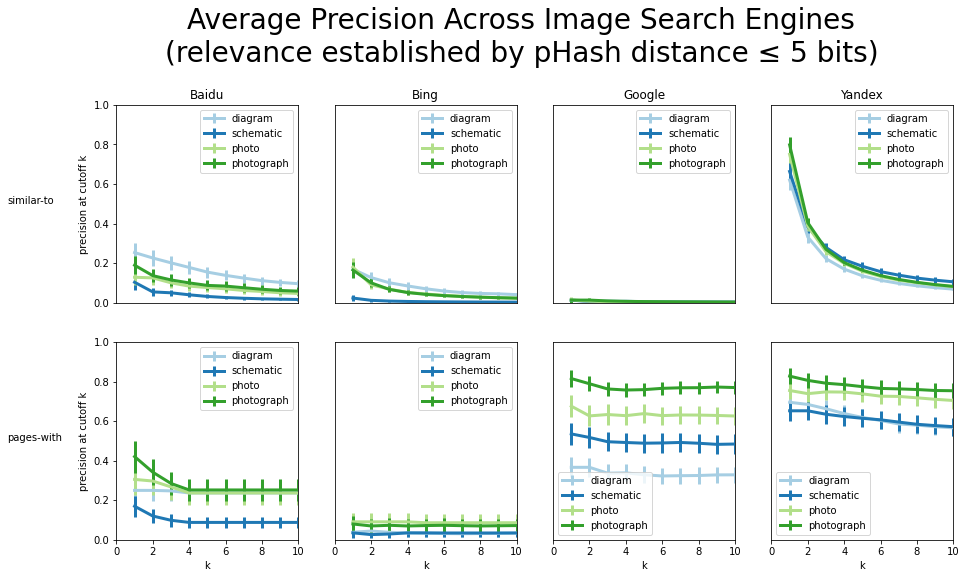}
    \caption{Average precision at cutoff $k$ between the values of 1 and 10 for Wikimedia Commons query images, using a pHash threshold of 5 bits. Error bars represent standard error at that value of $k$.}
    \label{fig:precision-phash}
\end{figure}

\begin{figure}
    \centering
    \includegraphics[width=\textwidth]{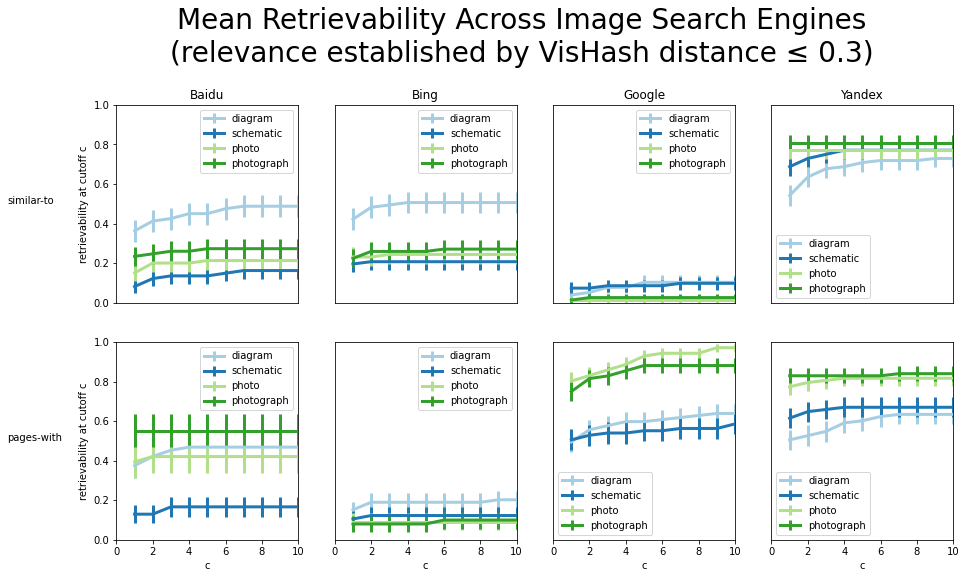}
    \caption{Mean retrievability at cutoff $c$ between the values of 1 and 10 for the Wikimedia Commons query images, using a VisHash threshold of 0.3. Error bars represent standard error at that value of $c$.}
    \label{fig:retrievability-vishash}
\end{figure}

\begin{figure}
    \centering
    \includegraphics[width=\textwidth]{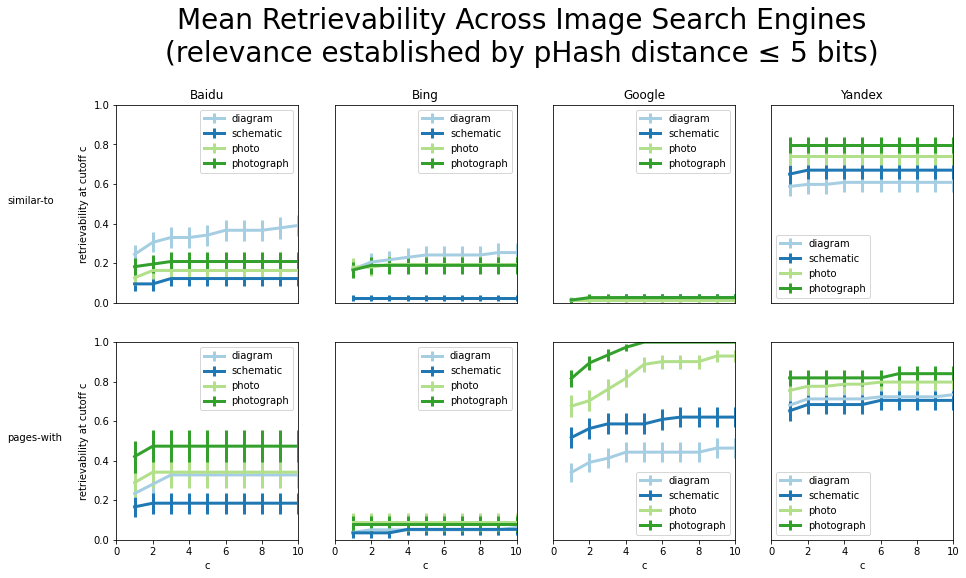}
    \caption{Mean retrievability at cutoff $c$ between the values of 1 and 10 for the Wikimedia Commons query images, using a pHash threshold of 5 bits. Error bars represent standard error at that value of $c$.}
    \label{fig:retrievability-phash}
\end{figure}

\begin{figure}
    \centering
    \begin{subfigure}{0.45\textwidth}
        \includegraphics[width=\textwidth]{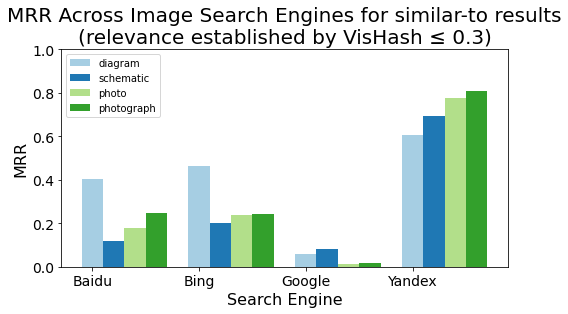}
        \caption{similar-to}
        \label{fig:mrr-vishash-similar-to}
    \end{subfigure}%
~
    \begin{subfigure}{0.45\textwidth}
        \includegraphics[width=\textwidth]{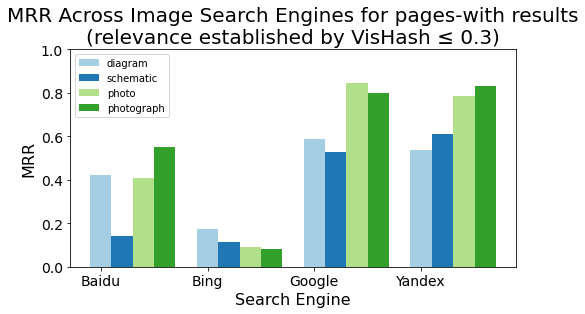}
        \caption{pages-with}
        \label{fig:mrr-vishash-pages-with}
    \end{subfigure}
    \caption{MRR scores for search engine results at VisHash distance $\le 0.3$}
    \label{fig:mrr-vishash}
\end{figure}

\begin{figure}
    \centering
    \begin{subfigure}{0.45\textwidth}
        \includegraphics[width=\textwidth]{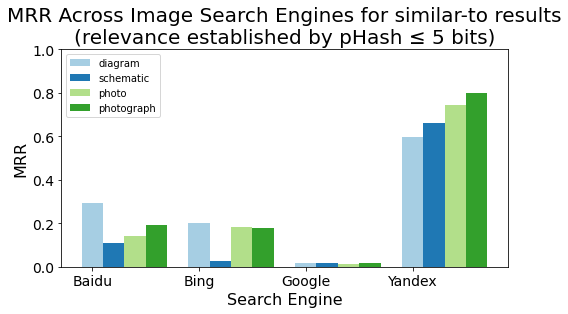}
        \caption{similar-to}
        \label{fig:mrr-phash-similar-to}
    \end{subfigure}%
~
    \begin{subfigure}{0.45\textwidth}
        \includegraphics[width=\textwidth]{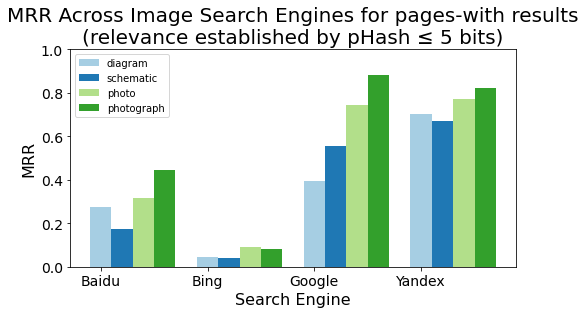}
        \caption{pages-with}
        \label{fig:mrr-phash-pages-with}
    \end{subfigure}
    \caption{MRR scores for search engine results at pHash distance $\le 5$ bits}
    \label{fig:mrr-phash}
\end{figure}

Tables \ref{tab:query-counts-similar-to} and \ref{tab:query-counts-pages-with} list the number of queries that returned results. These numbers are close to the total number of images per category (right column), giving us confidence in our scrapers' ability. In the rest of this section, we break down the results for the SERs with results.

Precision helps us understand how many relevant images were found within the cutoff $k$. Figure \ref{fig:precision-vishash} features the average precision scores for these queries with VisHash. Figure \ref{fig:precision-phash} shows the same for pHash. Green lines represent natural images in the categories of \emph{photo} and \emph{photograph}. Blue lines represent abstract images in the categories of \emph{diagram} and \emph{schematic}. We keep these colors consistent across all figures in this paper. We compute precision for each value of the cutoff $k$ within the first 10 SERs. A score of 1.0 is ideal. 

We see similar patterns with both image similarity measures (pHash and VisHash). For detecting if the same image exists in the result, recall that we selected a VisHash threshold $\le 0.3$ as recommended by Oyen et al. \cite{10.1117/12.2594720} and a pHash threshold of $\le 5$ bits as recommended by Krawetz \cite{krawetz_2011}. For similar-to results, we see that Baidu and Bing start with higher average precisions for \emph{diagram} than other categories, with VisHash showing higher scores. Google's precision is lower than 0.15 regardless of category and more severe with pHash than VisHash. Yandex starts strong with the highest precision. For pages-with results, the natural image categories score the highest precision across all search engines except for Bing. Google's precision scores are much higher with pages-with than with similar-to and are competitive with Yandex. For \emph{photograph} at $k=1$ and page-with results, Google scores a precision of 0.8158, which is much better than the 0.52 discovered by Bitirim et al. \cite{9202716} in 2020. Yandex performs slightly better than Google with \emph{photograph} at $k=1$ and pages-with results, scoring a precision of 0.8280. For Google, the difference in precision at $k=1$ and page-with results is greatest at 0.4491 between abstract images and natural images with \emph{diagram} at 0.3667 and \emph{photograph} at 0.8158.

Retrievability helps us determine if our query image was found within the cutoff $c$. Figure \ref{fig:retrievability-vishash} demonstrates the mean retrievability of images within the first ten results, as compared by VisHash. Figure \ref{fig:retrievability-phash} does the same for pHash. A score of 1.0 is ideal. 

For similar-to results, we see the highest mean retrievability scores with \emph{diagram} for Baidu and Bing. Google performs poorly here, regardless of the image similarity method. Yandex performs best but better for natural images compared to abstract ones. At c=1, Yandex experiences a maximum of 27\% difference in retrievability between \emph{photograph} (0.8085) and \emph{diagram} (0.5417) when measured with VisHash. This difference goes down to 19\% when measured with pHash.

When comparing the retrievability of pages-with results, for the most part, we see that all search engines tend to favor natural images over abstract ones. Bing has the poorest retrievability (less than 0.21) regardless of the image similarity method, so it is difficult to determine if it truly favors images in the category of \emph{diagram}. Yandex has high retrievability for natural images at $c=1$, but Google's retrievability passes Yandex by $c=10$. Google's and Yandex's retrievability is higher than the others for abstract images. At $c=1$, Yandex has a max 32\% difference in retrievability between \emph{photograph} (0.8297) and \emph{diagram} (0.5054) for pages-with results when measured with VisHash. At c=1, Google has a max 48\% difference in retrievability between \emph{photograph} (0.8157) and \emph{diagram} (0.3402) for pages-with results when measured with pHash. This number reaches reaches 54\% by $c=10$.

Our final measure, MRR, establishes how many results a visitor must review before finding the query image. A score of 1.0 is ideal because it indicates that the query image was found as the first result every time. Figure \ref{fig:mrr-vishash} shows a series of bar charts demonstrating the MRR scores established with VisHash. Figure \ref{fig:mrr-phash} shows the same with pHash. We represent each search engine with four bars, each bar representing an image category's performance. Again, green represents natural image categories while blue represents abstract images.

Yandex is the clear winner regardless of image similarity measure for similar-to results. Yandex consistently provides an MRR score of around 0.8 for natural images. Bing and Baidu are competitive for second place, and Google does not perform well for this result type. For pages-with results, Google performs competitively and scores slightly higher than Yandex in some cases. Baidu comes in a distant third, and Bing does not appear to be competitive.

Search engines crawl the web and add each page (and its images) to its \textbf{search index}. Maybe some of these search engines have not indexed the query image, which may explain the disparity in results. Unfortunately, it is difficult to assess the size of a search engine's index. Assuming that each search engine encountered these images via Wikipedia, we can search for the term ``Wikipedia'' in each search engine and discover how many pages the search engine has indexed with that term. Each page represents an opportunity to index an image from Wikimedia Commons. Using this method, the search engines, in order of result count, are Google (12.5 billion), Bing (339 million), Baidu (89.8 million), and Yandex (5 million). Assuming these numbers are an accurate estimate, then different index sizes do not explain the disparity in our results because Yandex, with the smallest number of indexed Wikipedia pages, still scores best by every measure.

Through our black-box analysis, by all measures, Google and Yandex perform best when returning pages-with image results, and they score best with natural images. Yandex performs best with similar-to image results and scores best with natural images. Google's results for similar-to may indicate that it is not trying to return the query image at all, instead favoring diversity over similarity.

\section{Future Work}

Here we report the preliminary results of a much larger experiment. We aim to build a stable dataset of Wikimedia Commons images that all four search engines have indexed; thus, we intend to repeat this experiment with more images. Once we have this list of images, we intend to conduct a follow-on study where we transform the images (e.g., cropping, resizing, rotating, color changes) to see how well the search engines perform despite each transformation.

\section{Conclusion}

Technical documents typically contain abstract images, but often users employ text queries to discover technical information. Computer vision has advanced such that we no longer need to only rely on text queries to retrieve images and find associated content. We can leverage the reverse image search capability of many general web search engines. Reverse image search allows one to upload an image as a query and review other images as search engine results. Few computer vision papers focus on retrieving abstract imagery; thus, we are uncertain if computer vision approaches that are successful with natural images perform as well with abstract ones. We leveraged the reverse image search capabilities of Baidu, Bing, Google, and Yandex to evaluate these black-box systems to answer the research question \emph{When using the reverse image search capability of general web search engines, are natural images more easily discovered than abstract ones?}

We experimented with 200 abstract images and 199 natural images from Wikimedia Commons. Using scrapers, we submitted these to the reverse image search engines of Baidu, Bing, Google, and Yandex to produce 154,191 results. Each search engine has two types of results: pages-with and similar-to. Pages-with results help the visitor discover which pages contain the same image as the one uploaded. Similar-to results help the visitor discover images that are like the one uploaded. Applying perceptual hashes to compare image queries to their SER images, we discovered that Yandex performs best in all cases through precision-at-k, retrievability, and mean reciprocal rank. Yandex also performs better for natural images than abstract ones. When considering pages-with results, Google, with a precision of 0.8191, and Yandex, with a precision of 0.8297, are competitive when reviewing the first result against its uploaded image. Both favor natural images in their performance, achieving a retrievability difference as high as 54\% between natural and abstract imagery at ten results.

Many reasons exist for users to conduct reverse image searches with abstract imagery. They may wish to protect intellectual property, build datasets, provide evidence for legal cases, establish scholarly evidence, or justify funding through image reuse in the community. That abstract images are at a disadvantage hurts users leveraging search engines for these use cases and provides opportunities for computer vision and information retrieval researchers.

\clearpage
%
%
\bibliographystyle{splncs04}
\bibliography{refs}
\end{document}